\documentclass[journal,twoside,web]{ieeecolor2}
\usepackage{generic}
\usepackage{cite}
\usepackage{amsmath,amssymb,amsfonts}
\usepackage{algorithmic}
\usepackage{graphicx}
\usepackage{textcomp}
\usepackage{caption}
\usepackage{graphicx}
\usepackage{amsmath,amssymb} 
\usepackage{color}
\usepackage{multirow}
\usepackage{mathrsfs}
\usepackage{subfigure}
\usepackage{bm}
\usepackage{array}
\usepackage{threeparttable}
\usepackage{float}
\usepackage{afterpage}
\usepackage{booktabs}
\def\BibTeX{{\rm B\kern-.05em{\sc i\kern-.025em b}\kern-.08em
    T\kern-.1667em\lower.7ex\hbox{E}\kern-.125emX}}
\begin{document}
	\title{Spatio-Temporal Progressive Attention Model for EEG Classification in Rapid Serial Visual Presentation Task}
	\author{Yang~Li,~\IEEEmembership{Member,~IEEE,} Wei Liu, Tianzhi Feng,
	Fu Li$^*$,~\IEEEmembership{Member,~IEEE,}
	 Chennan Wu, 
	Boxun Fu,
	Zhifu Zhao,~\IEEEmembership{Member,~IEEE,}
	Xiaotian Wang,~\IEEEmembership{Member,~IEEE,}
	Guangming Shi,~\IEEEmembership{Fellow,~IEEE}
	\thanks{Yang Li, Wei Liu, Tianzhi Feng, Fu Li, Chennan Wu, Boxun Fu, Zhifu Zhao, Xiaotian Wang and Guangming Shi are with the Key Laboratory of Intelligent Perception and Image Understanding of Ministry of Education, the School of Artificial Intelligence, Xidian University, Xi’an, 710071, China.\it{($^*$Corresponding author: Fu Li (E-mail: fuli@mail.xidian.edu.cn).)} \protect}
}
	\maketitle
	
	\begin{abstract}
		As a type of multi-dimensional sequential data, the spatial and temporal dependencies of electroencephalogram (EEG) signals should be further investigated. Thus, in this paper, we propose a novel spatial-temporal progressive attention model (STPAM) to improve EEG classification in rapid serial visual presentation (RSVP) tasks. STPAM first adopts three distinct spatial experts to learn the spatial topological information of brain regions progressively, which is used to minimize the interference of irrelevant brain regions. Concretely, the former expert filters out EEG electrodes in the relative brain regions to be used as prior knowledge for the next expert, ensuring that the subsequent experts gradually focus their attention on information from significant EEG electrodes. This process strengthens  the effect of the important brain regions. Then, based on the above-obtained feature sequence with spatial information, three temporal experts are adopted to capture the temporal dependence by progressively assigning attention to the crucial EEG slices. Except for the above EEG classification method, in this paper, we build a novel Infrared RSVP EEG Dataset (IRED) which is based on dim infrared images with small targets for the first time, and conduct extensive experiments on it. The results show that our STPAM can achieve better performance than all the compared methods.
		
	\end{abstract}


\section{Introduction}
\label{sec:introduction}
\IEEEPARstart{B}{rain-computer} interface (BCI) technology has made remarkable progress in recent years, such as communication aids~\cite{abiriComprehensiveReviewEEGbased2019} and prosthetic limbs~\cite{padfieldEEGBasedBrainComputerInterfaces2019}. Among the various BCI technologies, i.e., magnetoencephalography (MEG), EEG, and functional magnetic resonance imaging (fMRI)~\cite{clercReviewBrainComputer2013a}, EEG distinguishes itself due to its non-invasiveness and portability enabled by the ability to acquire data through electrode placement on the scalp~\cite{ramakuriPerformanceAnalysisHuman2019}, coupled with its high temporal resolution and cost-effectiveness. This makes EEG particularly suitable for various applications within BCI, including visual paradigms. Notably, EEG-based visual paradigms, such as steady-state visual evoked potentials (SSVEPs) and rapid serial visual presentation (RSVP), have proven effective in harnessing strong brain responses to visual stimuli~\cite{herrmannHumanEEGResponses2001}~\cite{10.1093/oxfordhb/9780195374148.001.0001}. 

In the paradigm of RSVP, image sequences are presented to subjects at a constant rate, generally 5-20 Hz, with the proportion of target and non-target images being approximately 10$\%$ and 90$\%$. These odd-ball target images can evoke distinct event-related potential (ERP) activities~\cite{POLICH20072128}. Particularly, the P300 component of ERP, characterized by a high amplitude response occurring at 250-500ms after target presentation, can offer a reliable marker for target detection in BCI applications~\cite{basar-erogluTopologicalDistributionOddball2001}~\cite{squiresEffectStimulusSequence1976}, which has garnered continuous attention in recent years~\cite{blackwoodCognitiveBrainPotentials1990}.

In RSVP-based BCI systems, the key to enhancing system performance is how to effectively classify the EEG signals, which interprets the generated activity of the brain in response to rapid serial visual presentations~\cite{liMultiTaskCollaborativeNetwork2024}. To this end, researchers have introduced various methods to enhance EEG signal decoding performance. For example, Sajda et al.~\cite{sajdaBlinkEyeSwitch2010} introduced HDCA, pioneering a dual-step approach combining FLD for spatial weight training with logistic regression for temporal weight learning. Blankertz et al.~\cite{blankertzSingletrialAnalysisClassification2011} introduced regularized linear discriminant analysis (rLDA) and further refined ERP signal classification, demonstrating the efficacy of appropriately regularized LDA via shrinkage. Li et al.~\cite{liAssemblingGlobalLocal2023} proposed an ensemble learning approach for EEG decoding in RSVP tasks, which is currently the best-performing conventional method. It leverages an extreme gradient boosting framework to sequentially generate sub-models, improving consistency with P300 patterns. 

Some researchers attempted to improve EEG signal classification performance by increasing the number of electrodes, but they found more electrodes may not lead to better performance~\cite{healyOptimisingNumberChannels2011}. Therefore, some researchers conducted plenty of experiments to find the best electrode combination but found the optimal varied among subjects~\cite{xu_multi-objective_2021}, so it is not feasible to select a uniform channel for all subjects~\cite{schroderRobustEEGChannel2005}.

In recent years, the landscape began to shift with the adoption of deep learning methods, which offered adaptive data representation learning and feature extraction, markedly improving RSVP EEG data classification~\cite{zhengweiwangReviewFeatureExtraction2018b}. Manor et al.~\cite{manorConvolutionalNeuralNetwork2015} introduced a novel spatio-temporal regularization via a tri-layered convolutional neural network. Further, realizing that the temporal resolution of EEG signals is much greater than that of spatial resolution, more and more research focuses on extracting the temporal features of EEG. Lan~\cite{lan_macro_2021} found there is information redundancy and interference noise in the time domain, they use a self-attention module to assign weights to each encoded slice. Li et al.~\cite{liPhasePreservationNeural2022} proposed PPNN to keep the phase information of EEG signals from being lost in convolution. Li et al.~\cite{liDecouplingRepresentationLearning2022b} further considered the class imbalance problem of RSVP tasks and proposed a DRL model to alleviate the negative impact. However, these methods are not yet sufficient to fully explore the temporal dependencies within the EEG signals.

Reflecting on the above literature, we identify three yet overlooked issues about RSVP EEG classification. The first one is which EEG electrodes contain more discriminative information. The current research has shown that the classification performance is seriously affected by the selected EEG electrodes in RSVP task~\cite{xu_multi-objective_2021}~\cite{lan_macro_2021}. These show the importance of EEG electrode selection. Thus, it is necessary to develop electrode selection methods for RSVP EEG classification. The second issue is how to utilize the time variation of RSVP EEG signals to improve the performance. As sequence data, the temporal dependencies inherent in EEG signals can provide valuable information that may be overlooked by traditional static EEG analysis methods. It is very valuable to explore which time period contains more discriminability. The third one is how to extend the application area of RSVP EEG decoding. Most of existed RSVP datasets are based on visible images with large targets. It is a challenge to investigate the performance of the RSVP paradigm based on other types of images.

To tackle the above issues, in this paper, we propose a novel attention model, denoted as STPAM, to progressively integrate the spatiotemporal information into the EEG features to boost the RSVP EEG classification. Specifically, STPAM consists of two major modules. The progressive spatial learning (PSL) module is designed to extract EEG features from more relative electrodes. To achieve this goal, we construct three spatial experts, where the former expert selects the important electrodes and feeds into the next one. This process continues to finely filter out the most ERP-related electrodes for RSVP EEG classification. After obtaining the spatial feature of all the EEG slices, the progressive temporal learning (PTL) module aims to select important EEG slices gradually and generate the spatial-temporal EEG features for final classification. Moreover, to increase the diversity of the above-selected results, we introduce divergence constraints to prompt the experts to select different electrodes and EEG slices. Besides, to address the third issue about the application of the RSVP task, we build a novel dataset IRED of twenty subjects' RSVP EEG data evoked by dim infrared images with small targets. The IRED is very helpful for exploring the application of the RSVP paradigm in low-light or nighttime environments and enriches the diversity of stimuli available for RSVP research. This dataset will be released to other researchers for further study.

The contributions of our work can be summarized as follows:

\begin{itemize}
	\item Propose a novel spatiotemporal EEG classification method, which is a progressively enhanced attention mechanism to select important electrodes and time slices;
	\item Build a new RSVP EEG dataset IRED in which the signals are evoked by dim infrared images with small targets. To the best of our knowledge, this is the first RSVP EEG dataset based on infrared images;
	\item The experiment results verified our approach's superiority in EEG classification tasks, evidencing state-of-the-art performance.
\end{itemize}

The remainder of this paper is organized as follows. Section~\ref{Sec: Dataset} introduces our IRED dataset based on infrared images. Section~\ref{Sec: The proposed method} details the our STPAM model. In Section~\ref{Sec: Experiment}, we present the experimental implementation and results from two datasets, followed by several experiments to discuss and evaluate our methods. Finally in Section~\ref{Sec: Conclusion}, we summarize the study.

\section{Construction of Infrared RSVP EEG Dataset and Proprocessing}
\label{Sec: Dataset}

\subsection{Subject}
The RSVP experiment involves twenty subjects (thirteen males and seven females, aged 19-25 years). All of them are students from Xidian University with normal or corrected-to-normal vision and no history of mental illness. The experiment adheres to the principles of the Declaration of Helsinki. All participants are briefed about the experimental process and provide their consent by signing the consent forms.
\subsection{Stimuli and Procedure}
\begin{figure}[htb]
	\includegraphics[width=1\columnwidth]{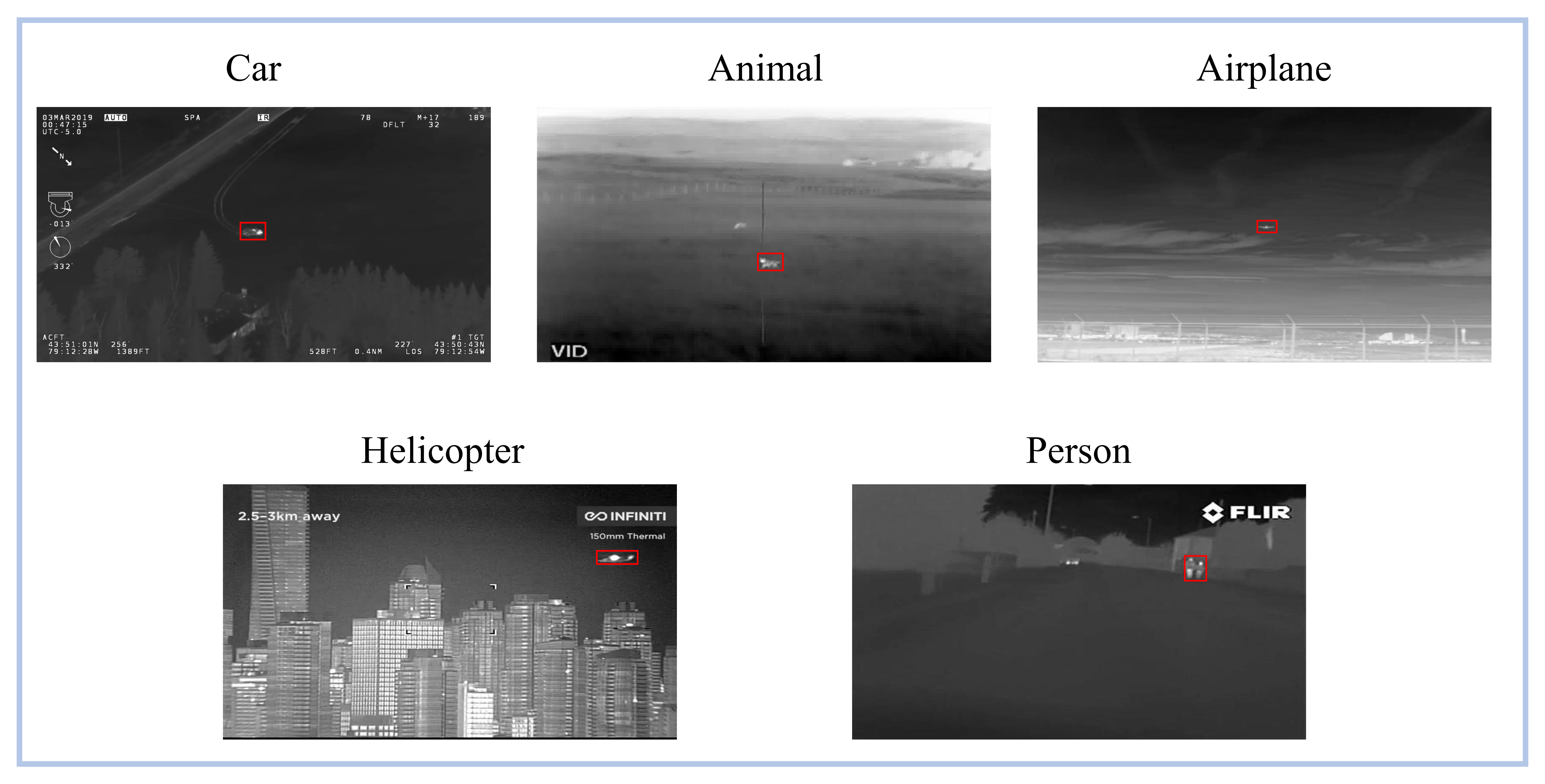} \\
	\caption{Examples of small target images. There are five types of targets in total including airplanes, animals, helicopters, cars, and persons.}
	\label{Fig: img}
\end{figure}
The experiment utilizes infrared images that come from the LSOTB-TIR dataset~\cite{liuLSOTBTIRLargeScaleHighDiversity2020}. First, we select the images with small targets and then resize them into 1280 $\times$ 720 pixels. As a result, we obtain 900 infrared images with small targets. Some examples are shown in Fig.~\ref{Fig: img}. These targets include airplanes, animals, helicopters, cars, and persons. 
\begin{figure}[htb]
	\includegraphics[width=1\columnwidth]{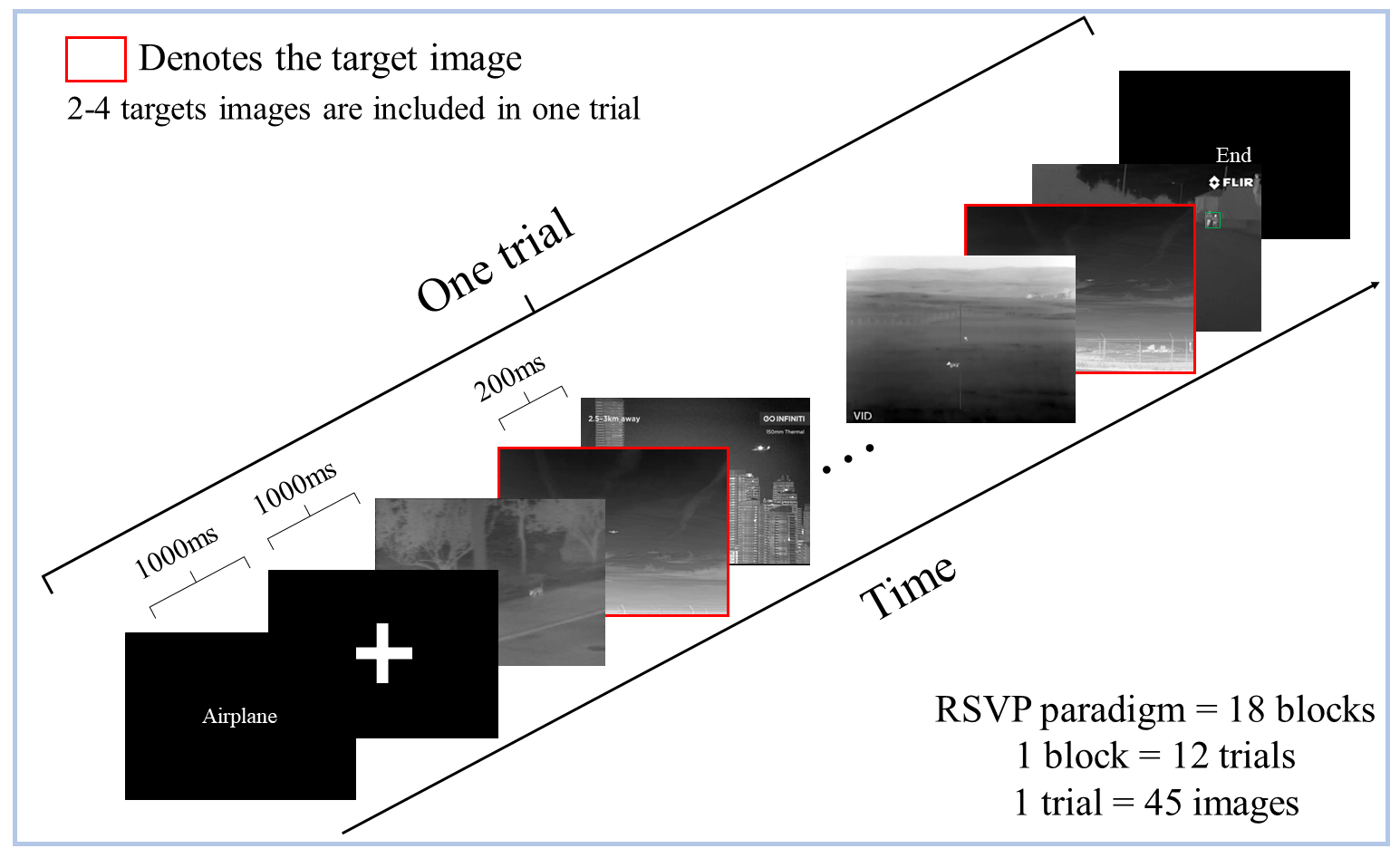} \\
	\caption{One trial in the RSVP paradigm. The target type of the current trial is shown for 1000ms first and then the fixation is presented for 1000ms to help the subjects to focus their attention on the center of the screen. Subquently each image is displayed for 200ms until the end.} 
	\label{Fig: paradigm}
\end{figure}
The overall RSVP paradigm is depicted in Fig.~\ref{Fig: paradigm}. First, a text is displayed on the screen to prompt the subject to focus on the occurrence of corresponding target images. Then a fixation appears for 1000 ms to ensure the subjects’ attention on the center of the screen. After that, the target and non-target images are displayed in sequence, which is called a trial. In each trial, 45 images that are displayed at a rate of 5 Hz. 2 to 4 of them are target images that occur at random positions in this trial to observe subjects' psychological expectations for the target images. To make this process clear, we highlight the target images with red boxes in Fig.~\ref{Fig: paradigm}. Note that these color cues are not presented to the subjects. Finally, the entire experiment comprises 18 blocks, with each block containing 12 trials. It is noteworthy that within each block, the remaining four types of images, excluding the one prompted as the target, will serve as the background images for the current block. Between every two blocks, the subjects are suggested to take a break when required. The paradigm code is implemented by Psychopy~\cite{peircePsychoPyPsychophysicsSoftware2007}.

\subsection{EEG Recording}
The EEG data were recorded using a BioSemi ActiveTwo system\footnote{https://www.biosemi.com/}, with 64 scalp electrodes based on the standard 10-20 system and a sampling rate of 1024 Hz. The impedance of each electrode was maintained below 20k$\Omega$ to ensure high-quality signals.

\subsection{Data Preprocessing}
The overall data preprocessing is listed as follows:
\begin{enumerate}
	\item[(1)] Sample segmentation. For the collected RSVP EEG signal, we first segment them into several EEG samples that start 1 second from the onset of each stimulus. Then we have about 1280 EEG samples for each subject. For each sample, its size is C$\times$ T, where C and T represent the number of EEG electrodes and time duration\footnote{For our collected EEG data, C and T are 64 and 256 respectively.}.
	\item[(2)] Filtering and standardization. After sample segmentation, all the samples are sent into a 6-order Butterworth band-pass filter with cut-off frequencies from 1 to 40 Hz and then are standardized using the z-score method.
\end{enumerate}

\section{The proposed model for RSVP EEG classification}
\label{Sec: The proposed method}
\begin{figure*}[htb]
	\centering
	\includegraphics[width=2.00\columnwidth]{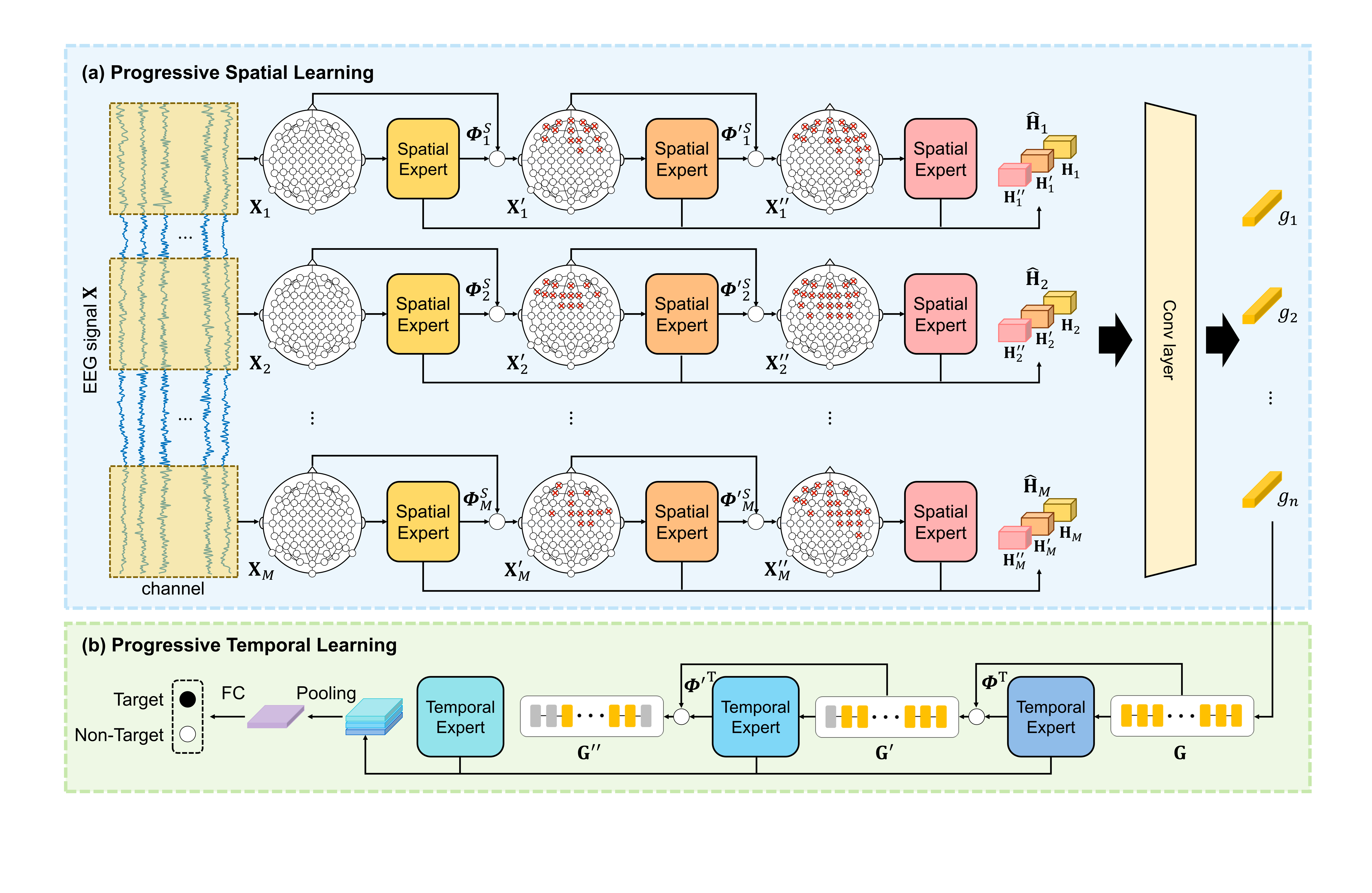} \\
	\caption{Framework of the proposed STPAM model. STPAM consists of two major blocks, i.e., progressive spatial and temporal learning blocks, which can pay attention to the important electrodes and time slices from spatial and temporal dimensions adaptively.}
	\label{Fig: model framework}
\end{figure*}

To specify the proposed method, we illustrate the framework of the STPAM model in Fig.~\ref{Fig: model framework}. STPAM consists of two major blocks, i.e., progressive spatial and temporal learning blocks, denoted as PSL and PTL, which can select the important electrodes and time slices from spatial and temporal dimensions adaptively. In STPAM, to capture the dynamic temporal information of EEG signals, we transform every EEG sample $\mathbf{X}\in \mathbb{R}^{C\times T}$  into another type $\mathbf{X}=\{\mathbf{X}_1, \mathbf{X}_2, \cdots, \mathbf{X}_M\}\in \mathbb{R}^{C\times T_M \times M}$ with a sliding window, in which $T_M$ and $M$ are the length of the window and the number of EEG
slices in one RSVP EEG sample respectively. The overall process is described as follows.

\subsection{Progressive Spatial Learning}
\label{Spatial}
\begin{figure}[htb]
	\centering
	\includegraphics[width=1\columnwidth]{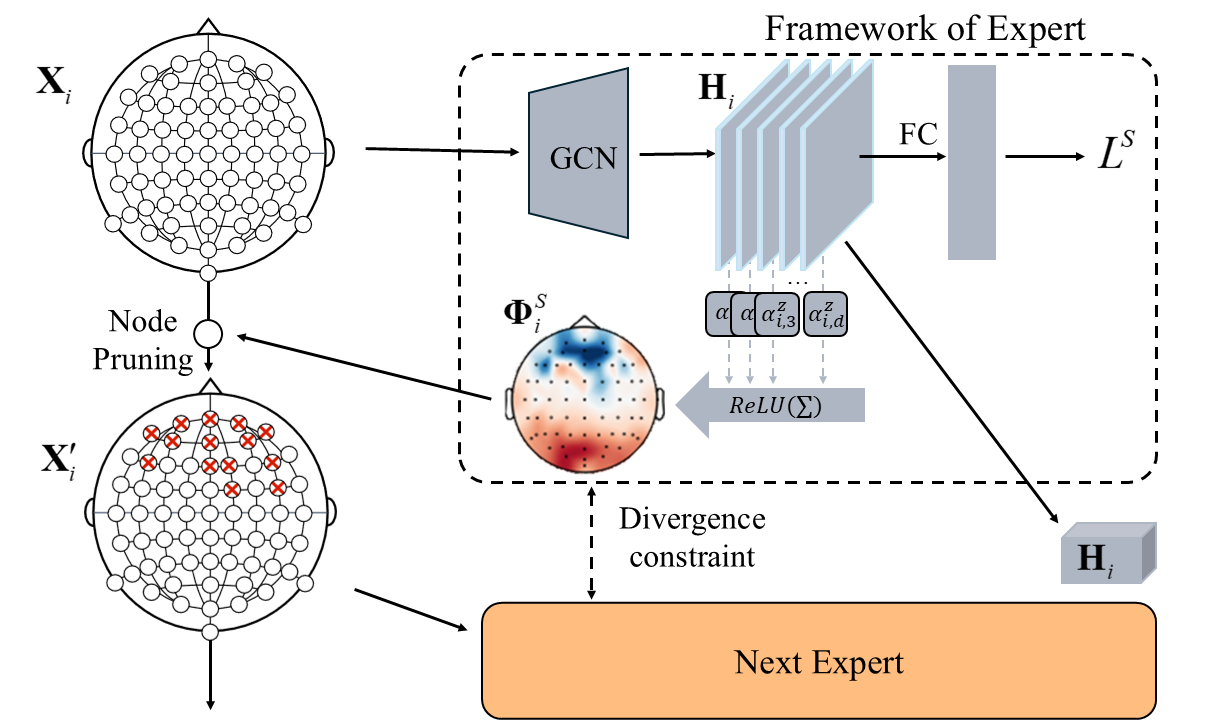} \\
	\caption{Structure of the spatial expert. }
	\label{Fig: expert}
\end{figure}
As well as extracting the spatial relationship, STPAM adopts three different experts to select less but more important electrodes progressively, which can capture the subtle spatial information based on different granularity.

For each expert in PSL, its structure is depicted in Fig~\ref{Fig: expert}. Concretely, for $i$-th EEG slice $\mathbf{X}_i=[\bm{x}_{i,1}, \bm{x}_{i,2}, \cdots, \bm{x}_{i,C}]^\top \in \mathbb{R}^{C\times T_M}$ in RSVP EEG sample $\mathbf{X}$, we construct an undirected graph $\mathcal{G}^S_i=(\mathcal{V},\mathcal{E})=(\mathbf{X}_i,\mathbf{A}^S_i)$, where $\mathcal{V}$ and $\mathcal{E}$ are the node and edge sets in the graph. To characterize the spatial relationship between the nodes in the graph, an adjacency matrix $\mathbf{A}^S_i\in \mathbb{R}^{C\times C}$ is utilized according to the positions of the electrodes. Here we employ the Chebyshev polynomials to approximate the graph convolution operation~\cite{defferrardConvolutionalNeuralNetworks2017} to extract high-level EEG features. For instance, supposed that $\mathbf{H}_i$ is the output of the first spatial expert for the $i$-th slice, the graph convolution can be formulated as:
\begin{eqnarray}
	\begin{split}
		\mathbf{H}_i&=&&\!\!\!\!\delta\left( \sum_{k=0}^{K}\omega^S_k T_k(\mathbf{Lap}^S)\mathcal{G}^S_i\right)\\
		&=&&\!\!\!\!\left[
		\begin{matrix}
			h_{i,11}, & h_{i,12}, &\cdots, &h_{i,1d}\\
			\vdots&\vdots &\vdots&\vdots\\
			h_{i,C1}, & h_{i,C2}, &\cdots, &h_{i,Cd}
		\end{matrix}
		\right]\in \mathbb{R}^{C\times d}, 
	\end{split}
	\label{GNN}
\end{eqnarray}
where $\delta(\cdot)$ is the activation function; $\omega^S_k$ represents the parameters to be learned in the convolution; $\mathbf{Lap}^S$ is the Laplace matrix after normalized; $d$ is the dimension after graph convolution and $T_k(\cdot)$ is the Chebyshev polynomial of order k. Based on features $\mathbf{H}_i,i\in\{1,2,\cdots,M\}$, we use a fully-connected layer to generate the probability distribution predictions of the experts. Consequently, the classification loss for the expert can be computed as:
\begin{eqnarray}
	&&{L}^S = -\frac{1}{N}\sum_{n=1}^{N}\tau(n,z)\log([\bm{{y}}^S_1(n)]_z),
	\label{loss1}\\
	&&\tau(n,z) = 
	\begin{cases}
		1, &\mbox{if } z = z_n^{label},\\
		0, &\mbox{otherwise},
	\end{cases}
	\label{loss2}\\
	&&\bm{{y}}^S_1(n) = \frac{1}{M}\sum^M_{i=1} f_1^S(\mathbf{X}_i(n)),
\end{eqnarray}
where $\mathbf{X}(n)$ represents the $n$-th RSVP EEG sample, $N$ is the amount of RSVP EEG samples. $z_n^{label}$ are the ground-truth label of the $n$-th sample. $f_1^S(\mathbf{X}_i(n))$ represents the probability distribution predicted by the first expert for the $i$-th slice of the $n$-th sample. $[\cdot]_z$ denotes the probability of the $z$-th class.

To transfer the knowledge obtained by the current expert to the next expert, we construct an attention map based on the gradient attention method~\cite{popeExplainabilityMethodsGraph2019}, which uses the gradient information to understand the importance of each node on the decision. To obtain the attention map, for the $j\in\{1,2,\cdots,d\}$-th row in feature $\mathbf{H}_i$, we first calculate the partial linearized slope from it for the class $z$. Then, a global average pooling is applied to calculate the $j$-th feature importance $\alpha_{i,j}^z$. This process can be represented as:
\begin{eqnarray}
	\alpha_{i,j}^z &=& \frac{1}{C}\sum^C_{c=1}\frac{\partial [f_1^S(\mathbf{X}_i)]_z} {\partial h_{i,cj}}.
	\label{attention map1}
\end{eqnarray}
Here $C$, $h_{i,cj}$ denote the number of electrodes and the value of the $c$-th row and $j$-th column in $\mathbf{H}_i$, respectively. Subsequently, we sum these feature maps based on the above weight $\alpha_{i,j}^z$ to obtain the heat-value $\phi^S_{i,c}$ of this electrode. Besides, we exclude electrodes with negative influence through the ReLU function. This process can be represented as: 
\begin{eqnarray}
	\phi^S_{i,c}&=&\sum_j \emph{ReLU}(\alpha_{i,j}^z h_{i,cj}).
	\label{attention map2}
\end{eqnarray}
Finally, for the $i$-th slice, we obtain the attention map $ \mathbf{\Phi}^S_i=\{\phi^S_{i,1}, \phi^S_{i,2}, \cdots, \phi^S_{i,C}\} \in \mathbb{R}^{C} $.  

Subsequently, according to the values in attention map $\mathbf{\Phi}_i^S$, STPAM identifies the more ERP-related electrodes with a threshold $\epsilon$. This process can be represented as:
\begin{eqnarray}
	\bm{x}'_{i,c}=
	\begin{cases}
		\bm{x}_{i,c}, &\mbox{if } \phi_{i,c}^S \ge \epsilon,\\
		0, &\mbox{if } \phi_{i,c}^S < \epsilon.
	\end{cases}
	\label{select electrode}
\end{eqnarray}
Consequently, we obtain the data $\mathbf{X}'_i = [\bm{x}'_{i,1}, \bm{x}'_{i,2}, \cdots, \bm{x}'_{i,C}]^\top \in \mathbb{R}^{C\times T_M}$ after selecting electrodes by former experts, excluding those irrelevant to classification and retaining those more important for classification. Note that considering the complexity of model training, here we keep the data size as $C \times T_M$ by zero padding.

To progressively focus on important electrodes, we construct three experts in the PSL module which can learn prior knowledge from the previous expert sequentially. These experts share the same structure as the above-described one but with different trainable parameters. In this case, as shown in Fig.~\ref{Fig: model framework}, the second expert can extract deeper EEG spatial feature $\mathbf{H}'_i\in \mathbb{R}^{C\times d}$ based on $\mathbf{X}'_i$, meanwhile producing a new attention map $\mathbf{\Phi}'^S_i$ and EEG data $\mathbf{X}''_i$ with more important electrodes. Similarly, based on $\mathbf{X}''_i$, the third expert can obtain the EEG spatial feature denoted as $\mathbf{H}''_i$. From the above three experts, we have obtained the spatial data representation of every slice as $\mathbf{\hat{H}}_i = [\mathbf{H}_i,\mathbf{H}'_i,\mathbf{H}''_i]\in \mathbb{R}^{(3C)\times d}$. Finally, to reduce the dimension of $\mathbf{\hat{H}}_i$, the PSL module performs a convolution and a fully connected layer to extract higher-level spatial feature as $\bm{g}_i \in \mathbb{R}^{d'}$, where $d'$ is the dimension after dimension reduction.
Herein, we have obtained the data representations learned from the PSL module that can be sequentially concatenated at each time slice and thus form a temporal sequence, which is denoted as $\mathbf{G}=[\bm{g}_1, \bm{g}_2, \cdots, \bm{g}_M]\in \mathbb{R}^{d'\times M}$.

\subsection{Progressive Temporal Learning}
\begin{figure}[htb]
	\centering
	\includegraphics[width=1\columnwidth]{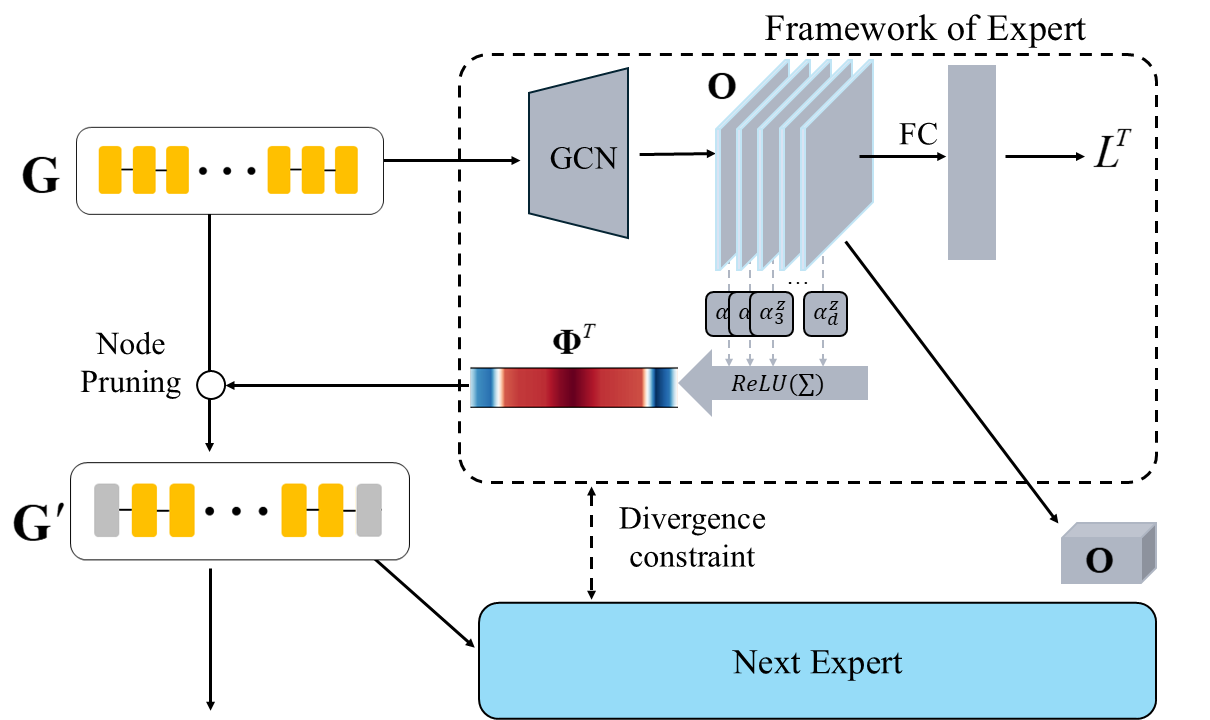} \\
	\caption{Structure of the temporal expert. }
	\label{Fig: timeexpert}
\end{figure}
As a nonstationary signal, the statistical properties of EEG are changing over time. Thus, the information in the time dimension is crucial for decoding RSVP EEG signals, and the contributions to the classification of EEG signals vary across different time periods. For this reason, we introduce three temporal experts to emphasize the contributions of different slices in an RSVP EEG sample, the structure is depicted in Fig.~\ref{Fig: timeexpert}.

Specifically, for the obtained feature sequence $\mathbf{G}=[\bm{g}_1, \bm{g}_2, \cdots, \bm{g}_M]\in \mathbb{R}^{d'\times M}$, we apply temporal graph convolution to learn the temporal information among this feature vector sequence. Concretely, STPAM treats each time slice $\bm{g}_i$ as a node and the chronological order as the edge to construct a temporal graph, which can be denoted as $\mathcal{G}^T=(\mathbf{G},\mathbf{A}^T)$. The result is expressed in Eq.~(\ref{GNN_time}) as
\begin{eqnarray}
	\begin{split}
		\mathbf{O}=\delta\left( \sum_{k=0}^{K}\omega^T_k T_k(\mathbf{Lap}^T)\mathcal{G}^T\right)\in \mathbb{R}^{d''\times M},
	\end{split}
	\label{GNN_time}
\end{eqnarray}
where $\omega^T_k$, $\mathbf{Lap}^T$, and $d''$ are the parameters of graph convolution, the normalized Laplace matrix, and the dimension after graph convolution, respectively. 

Subsequently, the classification loss of experts in the PTL module can be circulated as:
\begin{eqnarray}
	{L}^T = -\frac{1}{N}\sum_{n=1}^{N}\tau(n,z)\log([\bm{{y}}^T_1(n)]_z),\label{loss3}
\end{eqnarray}
where $\tau$ can be computed as Eq.~(\ref{loss2}), and the $\bm{{y}}^T_1(n)$ denotes the predicted distribution for the $n$-th sample by the first temporal expert.

Similar to the above PSL module, to transfer the knowledge among experts in the PTL module, refer to Eq.~(\ref{attention map1}) and (\ref{attention map2}) from the PSL to calculate the feature map importance $\alpha_j^z$ and then generate the attention map: 
\begin{eqnarray}
	\mathbf{\Phi}^T=\{\phi^T_1, \phi^T_2, \cdots, \phi^T_M\} \in \mathbb{R}^{M}.
\end{eqnarray}
According to $\mathbf{\Phi}^T$, referring to Eq.~(\ref{select electrode}), STPAM selects key RSVP EEG slices and disregards irrelevant and misleading parts, as the advanced knowledge passed from the previous expert to the next expert.

To progressively explore deeper temporal features with more prior knowledge, we construct three temporal experts in PTL modules. The corresponding outputs of the second temporal experts can be denoted as $\mathbf{O}'$. Between the second and the third temporal experts, the attention map $\mathbf{\Phi}'^T$ is generated by the former and guides the latter to generate its output $\mathbf{O}''$. Consequently, the final feature vector $\hat{\mathbf{O}}$ contains both the spatial and temporal information, which is expressed as
\begin{eqnarray}
	\hat{\mathbf{O}}=Pool[\mathbf{O},\mathbf{O}',\mathbf{O}''] \in \mathbb{R}^{d''\times M}.
\end{eqnarray}

Subsequently, we use a fully-connected layer and a softmax function to map these features into the category space. Supposed that $\tilde{\mathbf{o}}\in \mathbb{R}^{Md''}$ denote the vector that flattens $\hat{\mathbf{O}}$. The linear transformation in the fully connected layer can be formulated as:
\begin{eqnarray}
	\bm{e} = \tilde{\mathbf{o}}\mathbf{W} + \bm{b} \in \mathbb{R}^{Z},
\end{eqnarray}
where $Z$ represents the number of classes, and $e$ is the logits vector that will be input into the softmax function to produce probability distributions.

Consequently, the probability that RSVP EEG sample $\mathbf{X}$ is classified into the $z$-th class can be calculated as:
\begin{eqnarray}
	P(z|\mathbf{X})= \frac{\exp(e_z)}{\sum_{i=1}^{Z}\exp(e_i)},
\end{eqnarray}
where $e_i$ denotes value of the $i$-th class in $\bm{e}$.
Finally, the class corresponding to the highest probability is taken as the final prediction result:
\begin{eqnarray}
	\hat{z} = \arg\max_{z\in Z} P(z|\mathbf{X}).
\end{eqnarray}

\subsection{Optimization Objectives}
To improve the model's classification accuracy, we introduce a supervision term $L^c$ for the final prediction $\hat{z}$, which is a cross-entropy loss and can be calculated as follows:
\begin{eqnarray}
	L^c(\mathbf{X}|\theta_c) &=& -\frac{1}{N}\sum_{n=1}^{N}\log \Big(P(z_n^{label}|\mathbf{X}(n))\Big),
\end{eqnarray}
where $\theta_c$ denotes the parameters of the last fully connected layer, $z_n^{label}$ are the ground-truth label of the $n$-th sample.

Besides, we enhance the diversity of experts by introducing a divergence constraint among multiple attention maps. Our goal is to ensure that each expert's attention map is as distinct as possible from those of other experts. This constraint can be quantified using the Kullback-Leibler (KL) divergence, which is expressed as:
\begin{eqnarray}
	\begin{split}
		D_{KL}(\mathbf{\Phi}||\mathbf{\Phi}') =\sum_{j}^C[\phi_j\log(\phi_j)-\phi_j\log(\phi'_j)],						
	\end{split}
\end{eqnarray}
where C denotes the number of nodes, $\mathbf{\Phi}$ and $\mathbf{\Phi}'$ denotes two attention maps, $\phi_j$ and $\phi'_j$ represent the $j$-th value in $\mathbf{\Phi}$ and $\mathbf{\Phi}'$, respectively. For module PSL and PTL in STPAM, we calculate the losses of divergence constraint $L_{KL}^{\phi_s}$ and $L_{KL}^{\phi_t}$ to penalize the similarity of the attention maps. The losses can be computed as:
\begin{eqnarray}
	L^{KL}(\mathbf{X}|\theta_{\phi_S}) = \frac{1}{N}\sum_{n=1}^{N} \exp(-D_{KL}(\hat{\mathbf{\Phi}}^S_n||\hat{\mathbf{\Phi}}'^S_n)),\\
	L^{KL}(\mathbf{X}|\theta_{\phi_T}) = \frac{1}{N}\sum_{n=1}^{N} \exp(-D_{KL}({\mathbf{\Phi}}^T_n||{\mathbf{\Phi}}'^T_n)),
\end{eqnarray}
where $\theta_{\phi_T}$ and $\theta_{\phi_S}$ denote the parameters of GCN in PSL and PTL, respectively. Note that in the PSL module, each expert generates $M$ attention maps for $M$ slices, so we use $\hat{\mathbf{\Phi}}^S$ and $\hat{\mathbf{\Phi}}'^S$ to represent the average of these $M$ attention maps. Finally, the overall optimization objective of STPAM can be expressed as
\begin{eqnarray}
	\begin{split}
		\min\,&{L(\mathbf{X}|\theta_c, \theta_S, \theta_T, \theta_{\phi_S}, \theta_{\phi_T})} \\
		=&\,L^c(\mathbf{X}|\theta_c) + L^S(\mathbf{X}|\theta_S) + L^T(\mathbf{X}|\theta_T)   \\ + &\,\gamma \cdot \left( L^{KL}(\mathbf{X}|\theta_{\phi_S}) +  L^{KL}(\mathbf{X}|\theta_{\phi_T})\right).
	\end{split}
\end{eqnarray}
Here $L^c(\mathbf{X}|\theta_c)$ is responsible for enhancing feature extraction for final prediction. $L^S(\mathbf{X}|\theta_S) = \sum_{i=1}^{3}{L^S(\mathbf{X}|\theta_{S_i})}$ and $L^T(\mathbf{X}|\theta_T) = \sum_{i=1}^{3}{L^T(\mathbf{X}|\theta_{T_i})}$ is used to supervise the expert to learn more advanced knowledge from input and pass it on to the next expert, where $\theta_S$ and $\theta_T$ denotes the parameters to be learned in PSL and PTL modules, respectively.  $L^{KL}(\mathbf{\mathbf{X}}|\theta_{\phi_S})$ and $L^{KL}(\mathbf{X}|\theta_{\phi_T})$ penalize the similarity of the attention map, motivates the experts to generate attention maps that are distinct from those of preceding experts, thus increasing the diversity among experts. $\gamma$ is a hyper-parameter to adjust the proportion of the divergence loss in the overall loss.

\section{Experiments}
\label{Sec: Experiment}

\begin{table*}[htb]
	\caption{Classification Accuracy (\%) on Public Dataset}
	\centering
	\renewcommand{\arraystretch}{1.5}
	\tabcolsep=0.20cm
	\begin{threeparttable}		
		\begin{tabular}{cccccccc} 
			\toprule[1.2pt]
			\multirow{2}{*}{\textbf{Subject}} & \multicolumn{7}{c}{\textbf{Models}} \\ \cline{2-8}
			&rLDA~\cite{blankertzSingletrialAnalysisClassification2011}            
			&HDCA~\cite{sajdaBlinkEyeSwitch2010}          
			&EEGNet~\cite{lawhernEEGNetCompactConvolutional2018}	&EEG-Inception~\cite{santamaria-vazquezEEGInceptionNovelDeep2020}		&PPNN~\cite{liPhasePreservationNeural2022} 
			&XGB-DIM~\cite{liAssemblingGlobalLocal2023}
			&STPAM\\ \hline
			S01	&82.80	&75.58	&89.92	&91.47	&92.25	&90.70 &$\textbf{95.74}$\\ 
			S02	&78.81	&73.93	&86.32	&85.96	&90.60	&89.74 &$\textbf{92.74}$\\ 
			S03	&77.21	&73.91	&89.49	&91.30	&94.93	&91.67 &$\textbf{95.29}$\\ 
			S04	&85.61	&75.57	&89.69	&92.37	&94.66	&91.64 &$\textbf{95.06}$\\ 
			S05	&75.81	&74.76	&79.61	&82.50	&88.83	&90.49 &$\textbf{91.79}$\\ 
			S06	&70.25	&66.67	&77.70	&79.41	&82.48	&80.88 &$\textbf{85.29}$\\ 
			Average	&78.42 $\pm$ 4.93	&73.40 $\pm$ 3.09	&85.46	$\pm$ 4.99	&87.34 $\pm$ 4.79	 &90.63 $\pm$ 4.22	&89.79 $\pm$ 3.77	&$\textbf{92.65 $\pm$ 3.58}$\\ \bottomrule
			
		\end{tabular}
		\begin{tablenotes}[para]
			\footnotesize \hspace*{\fill} \\
		\end{tablenotes}
	\end{threeparttable}
	\label{Table: dataset1}
\end{table*}

\begin{table*}[htb]
	\caption{Classification Accuracy (\%) on IRED Dataset}
	\centering
	\renewcommand{\arraystretch}{1.5}
	\tabcolsep=0.25cm
	\begin{threeparttable}		
		\begin{tabular}{cccccccc} 
			\toprule[1.2pt]
			\multirow{2}{*}{\textbf{Subject}} & \multicolumn{7}{c}{\textbf{Models}} \\ \cline{2-8}
			&rLDA~\cite{blankertzSingletrialAnalysisClassification2011}            
			&HDCA~\cite{sajdaBlinkEyeSwitch2010}          
			&EEGNet~\cite{lawhernEEGNetCompactConvolutional2018}	&EEG-Inception~\cite{santamaria-vazquezEEGInceptionNovelDeep2020}		&PPNN~\cite{liPhasePreservationNeural2022} 
			&XGB-DIM~\cite{liAssemblingGlobalLocal2023}
			&STPAM\\ \hline
			S01	&70.62	&69.69	&75.31	&76.25	&76.56	&77.12	&$\textbf{77.19}$\\ 
			S02	&70.55	&65.95	&74.23	&73.01	&74.23	&72.09	&$\textbf{75.15}$\\  
			S03	&61.81	&68.48	&73.33	&73.33	&73.64	&73.64	&$\textbf{73.94}$\\ 
			S04	&69.55	&63.46	&76.60	&76.60	&77.24	&75.96	&$\textbf{77.56}$\\ 
			S05	&75.91	&70.73	&78.66	&78.96	&79.27	&79.27	&$\textbf{79.57}$\\ 
			S06	&61.65	&74.23	&75.46	&76.38	&75.46	&77.30	&$\textbf{78.53}$\\ 
			S07	&65.33	&67.48	&72.09	&73.93	&73.31	&72.09	&$\textbf{74.54}$\\ 
			S08	&68.26	&66.17	&75.75	&76.05	&76.35	&76.05	&$\textbf{77.25}$\\ 
			
			S09	&68.75	&72.81	&74.38	&75.94	&75.94	&76.25	&$\textbf{78.75}$\\ 
			S10	&72.29	&74.70	&79.52	&78.92	&80.42	&79.51	&$\textbf{81.33}$\\  
			S11	&76.06	&71.82	&77.88	&79.70	&79.70	&75.15	&$\textbf{80.91}$\\ 
			S12	&70.00	&74.55	&78.48	&78.48	&80.91	&79.09	&$\textbf{81.21}$\\ 
			S13	&69.09	&65.76	&71.52	&73.33	&73.03	&71.51	&$\textbf{74.55}$\\ 
			S14	&77.44	&80.18	&81.10	&81.71	&81.10	&83.53	&$\textbf{83.84}$\\ 
			S15	&73.01	&69.33	&76.07	&78.83	&77.91	&74.23	&$\textbf{79.75}$\\ 
			S16	&75.46	&72.09	&77.61	&78.22	&80.67	&76.68	&$\textbf{81.60}$\\ 
			S17	&83.64	&82.41	&85.19	&85.80	&87.35	&85.18	&$\textbf{87.96}$\\ 
			S18	&67.47	&65.96	&74.10	&72.59	&75.60	&71.69	&$\textbf{77.41}$\\ 
			S19	&70.25	&70.86	&74.23	&73.93	&77.61	&74.84	&$\textbf{78.22}$\\ 
			S20	&71.34	&72.56	&77.74	&76.83	&78.96	&73.17	&$\textbf{79.57}$\\
			Average&70.92 $\pm$ 5.06	&70.96 $\pm$ 4.69	&76.46 $\pm$ 3.13	&76.93 $\pm$ 3.20	&77.76 $\pm$ 3.35	&76.21 $\pm$ 3.63	&$\textbf{78.94 $\pm$ 3.29}$\\ \bottomrule
		\end{tabular}
		\begin{tablenotes}[para]
			\footnotesize \hspace*{\fill} \\ 
		\end{tablenotes}
	\end{threeparttable}
	\label{Table: dataset2}
\end{table*}

\subsection{Experimental setting}
\subsubsection{Dataset}
To evaluate the proposed method with the state-of-the-art methods, we conducted experiments on two RSVP EEG datasets in this paper, including our IRED and a public dataset. The details of IRED collected by ourselves are described in Section~\ref{Sec: Dataset}. Another public dataset for RSVP experiments is detailed as follows.

The public dataset contains RSVP EEG data from eight healthy volunteers (seven females and one male, aged 19-29 years)~\cite{rivetTheoreticalAnalysisXDAWN}, it is available at http://studycatalog.org/. The volunteers were required to wear a non-invasive EEG collection device called BioSemi ActiveTwo system, equipped with 256 scalp electrodes mounted on a full-head elastic electrode cap, with a sampling frequency of 256 Hz. During the data collection, the volunteers are required to sit in front of a CRT screen in a dimly lit environment for 256 trails, with each trial presenting a sequence of 50 images at a frequency of 12 Hz, two of which are target images containing 'airplanes'. The EEG signals from each trail were segmented into 1-second slices to serve as training samples, resulting in a 256${\times}$256 matrix. The samples underwent band-pass filtering with cut-off frequencies of 0.1-48 Hz. 

\subsubsection{Experimental protocol}
In this experiment, we use classification accuracy (acc) and standard deviation (std) as the criterion for comparing the performance of different methods. Specifically, for our IRED dataset with twenty healthy subjects' EEG data, there are approximately 1280 samples for each subject, consisting of target and non-target samples. To divide the training and test sets, 75$\%$ of the EEG data from each subject was selected as the training data, while the remaining 25$\%$ was designated as the test data. For the public dataset containing EEG data from eight subjects, we used data from the remaining six subjects for the subsequent experiments, as one subject provided only training data, and another subject's data was damaged. In the public dataset, the division into training and test sets was predefined during the collection phase. EEG data collected during the training session was used to train the models, while data from the testing session was used to evaluate classification performance~\cite{liPhasePreservationNeural2022}. 
\subsubsection{Implementation details}
Our model is trained on an NVIDIA GeForce RTX 2080Ti GPU, with CUDA 11.8 and cuDNN 8.6.0, within the PyTorch framework~\cite{paszkePyTorchImperativeStyle}. In the PSL module, the parameters are simply set as follows: $T_M$ = 32 and $M$ = 16. The Chebyshev polynomial order $K$=2, the number of filters $\omega^S$ is set to 32. Similarly, in the PTL module, the Chebyshev polynomial order $K$=3, the number of filters $\omega^T$ is set to 8.
For the attention module, we set the attention threshold $\epsilon$ to $0.2$.

We train our model using the cross-entropy loss function as the criterion, with the weight of the diversity constraint loss $\gamma$ simply set to 0.01. We use Adam~\cite{kingma2017adammethodstochasticoptimization} as our optimizer with a learning rate of 0.003. The source code is available on GitHub~\footnote{https://github.com/Mikufans0?tab=repositories}. 
\subsection{Experimental results}
\begin{table*}[htbp]
	\centering
	\captionsetup
	{justification=centering}
	\caption{\\Paired T-test Statistical Analysis Between Our STPAM and Other Methods on Two Dataset}
	\renewcommand{\arraystretch}{1.5}
	\tabcolsep=0.25cm
	\begin{threeparttable}		
		\begin{tabular}{ccccccc} 
			\toprule[1.2pt]
			\multirow{2}{*}{\textbf{Datasets}} & \multicolumn{6}{c}{\textbf{Models}} \\ \cline{2-7}
			&rLDA~\cite{blankertzSingletrialAnalysisClassification2011}            
			&HDCA~\cite{sajdaBlinkEyeSwitch2010}          
			&EEGNet~\cite{lawhernEEGNetCompactConvolutional2018}	&EEG-Inception~\cite{santamaria-vazquezEEGInceptionNovelDeep2020}		&PPNN~\cite{liPhasePreservationNeural2022}
			&XGB-DIM~\cite{liAssemblingGlobalLocal2023} \\ \hline
			Public dataset			&$\ddagger$	&$\ddagger$	&$\ddagger$	&$\dagger$	&*&$\dagger$	\\  
			IRED dataset	&$\ddagger$	&$\ddagger$	&$\ddagger$	&$\ddagger$	&$\ddagger$&$\ddagger$	\\ 
			\bottomrule
		\end{tabular}
		\begin{tablenotes}[para]
			\footnotesize
			\item[]Note:  *: $p\leq0.05$,  $\dagger: p\leq0.005$ , $\ddagger: p\leq0.001$
			\footnotesize \hspace*{\fill} \\
		\end{tablenotes}
	\end{threeparttable}
	\label{Table: compare}
\end{table*}
To evaluate the effectiveness of our proposed method, we conducted fair comparative experiments with six state-of-the-art methods in the relative field. These methods include:
\begin{enumerate}
	\item [(a)] rLDA~\cite{blankertzSingletrialAnalysisClassification2011}, which uses shrinkage estimators, demonstrates that appropriate regularization of linear discriminant analysis (LDA) by shrinkage yields better performance than other LDA-based methods. 
	\item [(b)] HDCA~\cite{sajdaBlinkEyeSwitch2010}, which is a two-stage conventional machine learning method for RSVP EEG classification.
	\item[(c)] EEGNet~\cite{lawhernEEGNetCompactConvolutional2018}, which is a compact convolutional neural network for EEG-based brain-computer interfaces, achieves comparably high performance across a range of BCI tasks.
	\item[(d)] EEG-Inception~\cite{santamaria-vazquezEEGInceptionNovelDeep2020}, which is the first model to use the Inception module for ERP detection.
	\item[(e)] PPNN~\cite{liPhasePreservationNeural2022}, which is a network that learns phase information to improve the EEG classification in RSVP tasks, representing the classic method for RSVP tasks.
	\item[(f)] XGB-DIM~\cite{liAssemblingGlobalLocal2023},  which is an ensemble learning-based algorithm XGB-DIM that adopts one global spatial-temporal filter and a group of local filters to extract discriminant information.
\end{enumerate}

The experimental results of the above two datasets are summarized in Table~\ref{Table: dataset1},~\ref{Table: dataset2}. 

From Table~\ref{Table: dataset1}, we can observe that our proposed model outperforms all compared methods on the public RSVP dataset. STPAM achieves the highest classification accuracy among the six subjects. Compared with other methods, the average accuracy is improved by 14.23$\%$, 19.25$\%$, 7.19$\%$, 5.31$\%$, 2.02$\%$, and 2.86$\%$, respectively. Additionally, our method maintains a low standard deviation, demonstrating the stability of STPAM in classification tasks. This improvement can be attributed to STPAM's enhanced ability to extract spatial information from the 256 scalp electrodes. This dataset contains 256-channel EEG data, providing abundant spatial information for EEG decoding, but also introducing additional interference. Previous methods, such as PPNN, EEGInception, and EEGNet, struggle to effectively extract spatial information, which consequently limits their performance. In contrast, our progressive adaptive electrode selection method, combined with graphical strategies, enables the model to maximize the useful information from these electrodes while minimizing the irrelevant information as much as possible.

From Table~\ref{Table: dataset2}, we observe that the proposed model achieves superior performance on the IRED dataset than all other methods across all subjects. The improvement are 8.01$\%$, 7.98$\%$, 2.47$\%$, 2.00$\%$, 1.17$\%$, 2.74$\%$, respectively. Additionally, a comparison of the results presented in Table~\ref{Table: dataset1} and Table~\ref{Table: dataset2} reveals pronounced differences in accuracy between the two datasets. This decline in IRED can be attributed to the unique characteristics of the IRED dataset, which utilizes infrared images with small targets as stimuli. These images make it challenging  to immediately detect the target when the target image is presented~\cite{fan_dc-tcnn_2022}. Therefore, the timing of ERP appearance may vary significantly across different EEG samples, posing a substantial challenge to current RSVP EEG classification methods. In contrast to EEGInception, which leverages multi-scale features, and PPNN, which incorporates phase information, SPTAM minimizes the impact of ERP appearance time differences on classification by slicing EEG data and employing progressive attention to selectively focus on critical time slices. Consequently, it delivers superior performance on the IRED dataset.

To test if the proposed STPAM is statistically significantly superior to the other six methods, we conducted paired t-test analyses and the results are shown in Table~\ref{Table: compare}. Each row lists the results of the statistical comparison between STPAM and other models on the public dataset and the IRED dataset. Note that different symbol represents different significance levels and the $p$-value below the conventional threshold of 0.05 is considered indicative of statistically significant improvement. Specifically, even when compared to the latest methods PPNN and XGB-DIM, STPAM maintains statistical significance ($p < 0.05$) across both datasets. Moreover, on the more challenging IRED dataset, STPAM achieves highly significant differences ($p < 0.001$) against all other methods. These results demonstrate that our approach exhibits statistically significant improvements in overall comparison methods, with particularly on the IRED dataset.

\subsection{Discussion}
\label{Sec: Discussion}

\begin{figure*}[htb]
	\includegraphics[width=2.02\columnwidth]{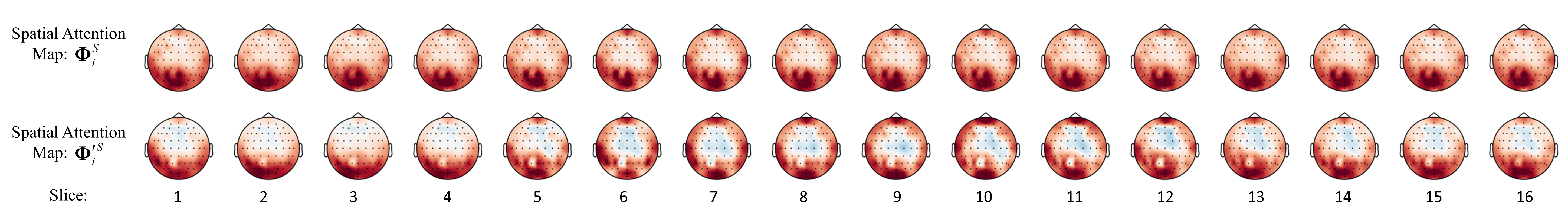} \\
	\caption{The average of attention maps that spatial experts generated for all subjects on the IRED dataset. Electrodes deemed more important by the spatial expert are represented with deeper red.}
	\label{Fig: Spatial map}
\end{figure*}

\begin{figure*}[htb]
	\includegraphics[width=2.02\columnwidth]{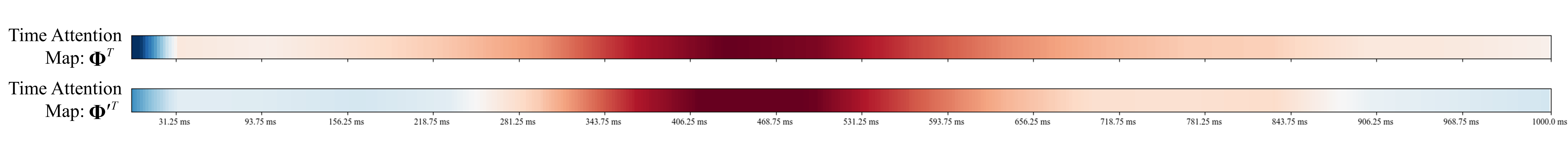} \\
	\caption{The average of attention maps that temporal experts generated for all subjects on the IRED dataset. Time periods deemed more important by the temporal expert are represented with deeper red.}
	\label{Fig: Time map}
\end{figure*}
\subsubsection{Performance based on different numbers of experts}
\label{Sec: Number of Expert}

The better classification performance of STPAM may largely be attributed to the progressive attention mechanism implemented through the multiple-experts approach. To validate this point, we conducted further experiments based on a simplified model obtained by removing certain experts.
\begin{enumerate}
	\item The spatial-temporal model (STM), removes two experts in both the PSL and PTL modules, thereby deleting the attention mechanism between the experts.
	\item The Spatial-Temporal Attention Model (STAM), removes one expert in both the PSL and PTL modules, retaining one attention between the first two experts.
\end{enumerate}
\begin{table}[htbp]
	\centering
	\captionsetup{justification=centering}
	\caption{Classification Accuracy (\%) of Models with Different Numbers of Experts on Two Datasets}
	\renewcommand{\arraystretch}{1.5}
	\tabcolsep=0.45cm
	\begin{threeparttable}		
		\begin{tabular}{cccc}
			\toprule[1.2pt]
			\multirow{2}{*}{\textbf{Dataset}} & \multicolumn{3}{c}{\textbf{Models}} \\ \cline{2-4} 
			& STM & STAM & STPAM	\\ \hline
			Public Dataset	& 88.41 & 90.20 & $\textbf{92.65}$\\ 
			IRED Dataset	& 76.64 & 77.33 & $\textbf{78.94}$\\ \bottomrule
		\end{tabular}
		\begin{tablenotes}[para]
			\footnotesize 
		\end{tablenotes}
	\end{threeparttable}
	\label{Table: Expert}
\end{table}

The comparative results are displayed in Table~\ref{Table: Expert}. Specifically, on the public dataset, the STM model achieved a classification accuracy of 88.41\%, while the STAM model improved this to 90.20\%, with an increase of 1.79\%. The STPAM model, which employs a progressive attention mechanism, reached 92.65\%, further improving by 2.45\% compared with STAM and 4.24\% compared to STM. On the IRED dataset, STM obtains an accuracy of 76.64\%, which is raised to 77.33\% by STAM, with an increase of 0.69\%, and further improved to 78.94\% by the STPAM model, showing an overall increase of 2.30\% over STM and 1.60\% over STAM. Besides, compared with other models in Table~\ref{Table: dataset1} and \ref{Table: dataset2}, STM achieves comparable performance to EEGNet because of the ability to capture the relationship information between different electrodes and time slices. Moreover, by integrating an attention mechanism to selectively focus on significant electrodes and time slices, STAM surpassed the performance of STM, achieving results that are on par with those of PPNN and XGB-DIM. Furthermore, STPAM, which introduces a progressive attention mechanism that leverages prior knowledge learned from previous experts to enhance subsequent experts, achieves the highest performance. Overall, the introduction of attention mechanisms significantly improved the classification performance of the models, and the progressive attention mechanism demonstrated even greater advantages.

\subsubsection{Visualization of attention mechanism}
\label{Attention maps}

To visualize the impact of the different electrodes on RSVP EEG classification, the attention map $\mathbf{\Phi}^S_i$ and $\mathbf{\Phi}'^S_i$ are presented in the first and second rows of Fig.~\ref{Fig: Spatial map}~\footnote{It is generated by MNE-Python~\cite{gramfort_mne_2014}.}, respectively. The first row indicates that posterior parietal and occipital regions make substantial contributions to classification. This is because the parietal lobe plays a crucial role in combining visual and somatosensory data~\cite{lynchParietalLobeMechanisms1977}, while the primary function of the occipital lobe is the processing of visual inputs~\cite{PMID:10221426}. 
The attention map of the second row indicates that more irrelevant electrodes were excluded, while the posterior parietal and occipital regions remained the focus of attention. Additionally, the temporal lobe made significant contributions to classification during certain time periods, which might be attributed to the dim targets in the IRED dataset further activating the temporal lobe~\cite{knutsonVisualDiscriminationPerformance2012a}. This indicates that the prior knowledge has provided valuable electrode information for subsequent experts.

Besides, to visually demonstrate the varying contributions of different time slices in RSVP EEG classification, we depicted the temporal attention maps $\mathbf{\Phi}^T$ and $\mathbf{\Phi}'^T$ in the first row and the second row of Fig.~\ref{Fig: Time map}, respectively. It can be observed that the time period from approximately 150ms to 800ms contributes significantly to the classification, as most of the event-related potentials occur within this time frame~\cite{10.1093/oxfordhb/9780195374148.001.0001}. Further observation reveals that the 350ms to 550ms interval is considered the most critical period, which is slightly later than the typical onset of the P300 wave. This delay is attributed to the fact that our IRED dataset uses small-target infrared image stimuli, which require subjects to take some time to detect the targets within these images. This further shows that our method is effective in identifying more critical time slices.

\subsubsection{Visualization of spatial-temporal features}
\label{TSNE}
\begin{figure}[htb]
	\centering
	\includegraphics[width=1\columnwidth]{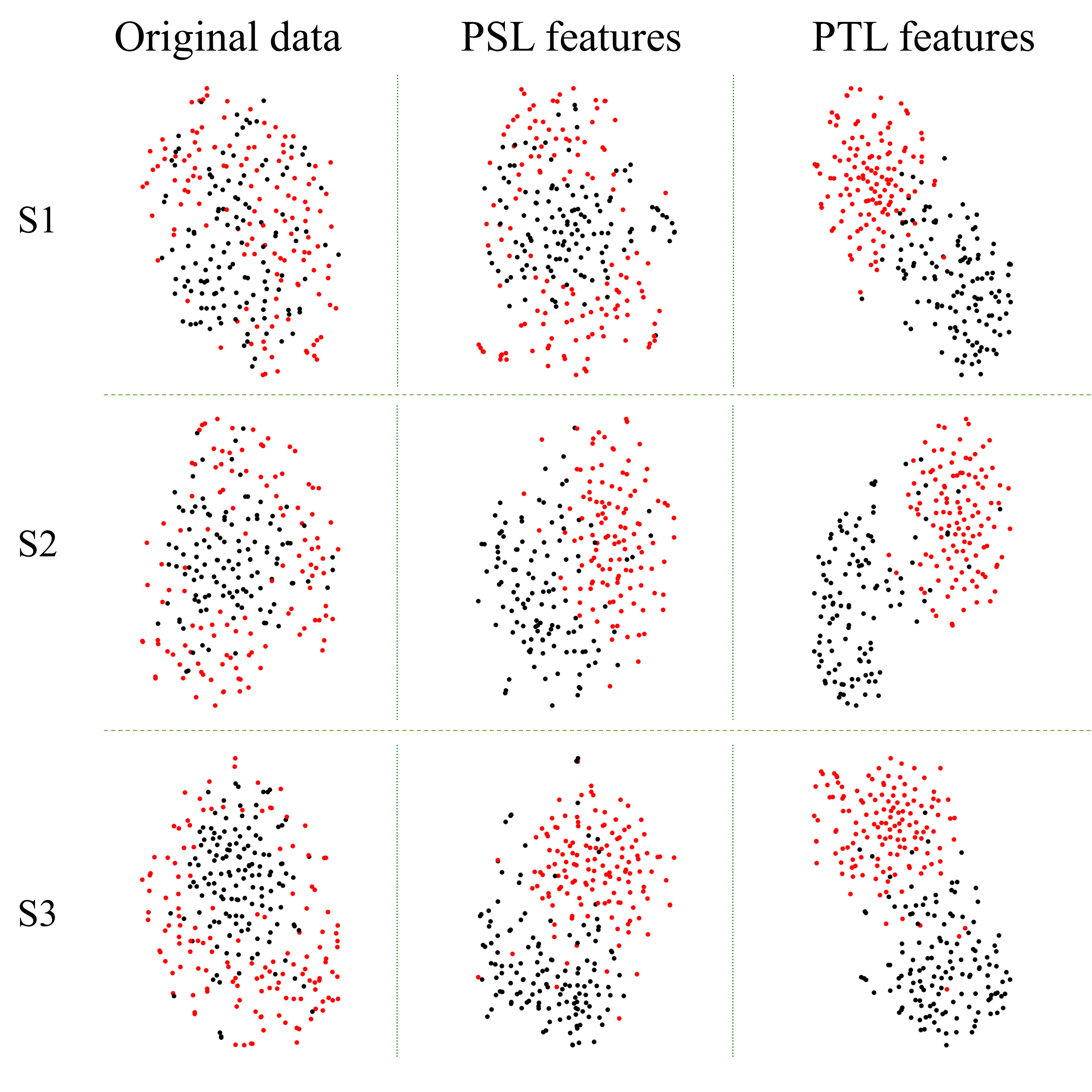}
	\caption{t-SNE visualization. Red dots denote target samples, whereas black dots represent non-target samples.}
	\label{Fig: tsne}
\end{figure}

In the previous experiment, we validated the effectiveness of the proposed spatial and temporal progressive attention. To further verify this, we visualize the feature distributions associated with three subjects from the public dataset using the t-distributed stochastic neighbor embedding (t-SNE)~\cite{JMLR:v9:vandermaaten08a} technique, which is extracted by the PSL and PTL module. The results are shown in Fig.~\ref{Fig: tsne}. The figure shows that the red and black dots are intermixed in the original data, indicating weak separability. In contrast to the raw features, we can observe that the scatter plot shows some degree of separation between the red and black dots after extracting by the PSL module. Furthermore, compared with the PSL module, we observe that the two classes of dots are separated, and dots of the same class are closer to each other after PTL captures more discriminative temporal features.

\section{Conclusion}
\label{Sec: Conclusion}
In this study, we introduce a deep learning model designed for RSVP classification, addressing challenges in existing methods by enhancing spatiotemporal connectivity and optimizing the use of RSVP-relevant information. By incorporating a graph-based structure and progressive attention mechanisms, our model can refine the processing and interpretation of EEG data for better classification. Besides, based on dim infrared images with small targets, we collect an RSVP EEG Dataset called IRED, which will significantly aid in extending the usage scenarios for RSVP. Evaluations of two datasets demonstrate that our model outperforms several established benchmarks, highlighting its potential in handling complex EEG signal characteristics.

\newpage
\bibliographystyle{IEEEtran}
\bibliography{STPAM}

\end{document}